\documentclass[lettersize,journal]{IEEEtran}
\usepackage{amsmath,amsfonts}
\usepackage{amssymb}
\usepackage{algorithmic}
\usepackage{algorithm}
\usepackage{array}
\usepackage[caption=false,font=normalsize,labelfont=sf,textfont=sf]{subfig}
\usepackage{textcomp}
\usepackage{stfloats}
\usepackage{url}
\usepackage{verbatim}
\usepackage{graphicx}
\usepackage{cite}
\usepackage{booktabs}
\begin{document}

\title{Learning by Examples: Power Grid Steady State Analysis with a Tabular Foundation Model}

% \author{Olayiwola Arowolo, Jochen Cremer, Peter Palensky, Oluwadara Adedeji ~\IEEEmembership{Electrical and Sustainable Energy, TU Delft}
        % <-this % stops a space

\author{\IEEEauthorblockN{Olayiwola Arowolo$^{*}$, Oluwadara Adedeji$^{\ddagger}$, Peter Palensky$^{*}$, Jochen L. Cremer$^{*}$$^{\dagger}$}\\
\IEEEauthorblockA{$^{*}$Delft University of Technology, Delft, the Netherlands \\
$^{\dagger}$Austrian Institute of Technology, Vienna, Austria \\
$^{\ddagger}$University College Dublin, Belfield Dublin, Ireland \\
o.a.arowolo, j.l.cremer, p.palensky\{@tudelft.nl\}, oluwadara.adedeji\{@ucdconnect.ie\}\\
}
}        

\markboth{Journal of \LaTeX\ Class Files,~Vol.~14, No.~8, August~2021}%
{Shell \MakeLowercase{\textit{et al.}}: A Sample Article Using IEEEtran.cls for IEEE Journals}


\maketitle

\begin{abstract}
Foundation models have shown strong generalization across scientific domains, but their use in power systems remains largely task-specific. We investigate whether a pretrained tabular foundation model (TFM), TabPFN, can serve as a general-purpose neural solver for power grid steady-state analysis. We evaluate the TFM on three tasks: power flow (PF) prediction, N-k contingency screening (NCS), and voltage stability margin (VSM) prediction. We performed experiments on 118- and 2000-bus systems, comparing the TFM against task-specific neural baselines. Without retraining or fine-tuning, the TFM matches or outperforms the best task-specific model in several settings across the three tasks. For 118-bus PF prediction, a variant of the TFM achieves errors of $1\times 10^{-3}$ pu for voltage magnitudes and $0.30^\circ$ for voltage angles, outperforming all baselines. On unseen N-1 contingency and close-to-infeasible PF cases, the TFM is competitive with the DCPF approximation. For millions of unseen N-2 contingencies, the TFM achieves AUCs of 0.92 and 0.95 for branch and voltage violation screening, respectively. The TFM also attains accurate VSM prediction performance for the 2000-bus system with limited labelled data. The TFM attains strong results across the tested tasks despite pretraining without any power-system-specific dataset and requiring no fine-tuning step. These results offer a glimpse into the potential of pretrained TFMs as a neural solver for a range of power grid analysis tasks.

\end{abstract}


\section{Introduction}
Driven by massive integration of renewable energy and rapid electrification of mobility and industry, the global energy transition introduces significant variability and uncertainty into grid operations \cite{kroposki_achieving_2017}. Therefore, grid operators must rapidly assess grid security and reliability while maintaining adequate safety margins in the constantly changing operational landscape \cite{bellizio_machine-learned_2022}. Power flow (PF) is the core task in many steady-state security analyses, including contingency screening, voltage stability analysis, and topology optimization \cite{rivera_pf_2025}. Classical AC PF solvers provide accurate solutions but become computationally expensive at large scales. The computational cost becomes especially burdensome for applications which require repeated PF solver calls. \cite{hamann_foundation_2024}. For example, exhaustive N-3 branch contingency screening on the modest IEEE-118 bus system requires over 1 million PF solver calls per operating point (OP). Operators typically rely on DC power flow approximation (DCPF) for such applications, but DCPF has large errors for high loading conditions \cite{simpson-porco_lossy_2018}. These errors lead to conservative operation of the power system and increases the risks of blackouts. Therefore, conventional computational tools for planning and operating the grid are increasingly inadequate for real-time monitoring and analysis across multiple scenarios.

Machine learning (ML) models have been proposed as capable neural surrogates for many power system analyses. Despite the promising speed advantages offered by ML, its adoption in practical power system operation remains limited by several factors. These include human barriers such as lack of trust in black-box models \cite{chatzivasileiadis_machine_2022} and technical barriers such as scarcity of large, high-quality datasets \cite{gillioz_large_2025}. Limited generalization is also an important technical barrier that current ML approaches face in power systems. We define generalization as the ability of an ML model to give accurate predictions for OPs outside its training data \cite{arowolo_exploring_2025}, where the OP includes the loading, injection, and topology of the grid. Since it is practically impossible to cover all possible topology and parameter combinations in the training data of large grids, many ML models have to be retrained when an unseen topology/parameter configuration is encountered during operation \cite{bellizio_machine-learned_2022}. Transfer learning in power systems \cite{ren_transfer_2020} can suffer from negative transfer and catastrophic forgetting \cite{kirkpatrick_overcoming_2017}. Separate ML models may also be trained and interfaced for individual components in the grids \cite{stiasny_pinnsim_2024}. However, these approaches are not scalable.

Graph Neural Networks (GNNs) have been explored as a way to improve the generalization of ML models to topological changes in power grids because of their inherent generalization properties \cite{gama_stability_2020}. GNNs have demonstrated promising generalization to N-1 topology changes on tasks such as PF \cite{hansen_power_2023} and ACOPF \cite{arowolo_towards_2026}. However, GNNs face several architectural limitations such as limited expressivity, oversmoothing, and oversquashing \cite{alon_bottleneck_2021}. These limitations can cause poor performance of GNN models in applications with little or no locality dependence compared to topology-agnostic models \cite{falconer_leveraging_2023}. While different modifications have been proposed to overcome these limitations in isolation, developing a model that combines topology flexibility and sufficient accuracy across a range of power system tasks remains challenging. Existing GNN approaches also remain largely task- and grid-specific. Different models are typically trained for tasks which have strong dependence on each other, for example, PF and N-k contingency screening, which highlights the limited generalization capability of existing GNN approaches.

A  foundation model is a large-scale transformer-based model which is pretrained on large, diverse datasets through self-supervision and learns transferable knowledge that generalizes to new, unseen tasks \cite{hamann_foundation_2024}. FMs have recently emerged as a path towards better generalization of ML models in power systems and have already demonstrated success in domains such as language, vision, and audio. Application-specific FMs have been proposed in power systems.\cite{yu_pricefm_2025} proposed time series FMs for electricity price data, \cite{cremer_customising_2026, huang_large_2024} investigated integrating existing LLMs into power system workflows, while \cite{li_predicting_2026} proposed a foundation model for power system dynamic simulations. While these approaches extend the generalization of ML to specific applications through custom-made FMs or existing large language models (LLMs), they still fall short of a general-purpose FM model which can be applied to a wide range of power system tasks.
\cite{hamann_foundation_2024} discussed the idea of a multi-task grid foundation model (GridFM), while \cite{puech_genco_2026} proposed a GNN architecture towards a GridFM. While promising, these approaches apply to a limited set of tasks covered in the pretraining. 

In this work, we investigate tabular foundation model (TFM) based on prior-fitted networks \cite{muller2022transformers} as a promising general-purpose GridFM. TFMs are pretrained transformer models designed for making predictions on tabular datasets. These models have demonstrated impressive data efficiency and accuracy across a wide range of tasks \cite{hollmann_accurate_2025}. Unlike classic ML, TFM works by in-context learning (ICL). That is, the labelled training data serves as examples/contexts for the model during inference. This is similar to how large language models infer responses from context provided in the problem. For a power grid, this means the same model can be used to predict for a new task, distributions of OP or grid topologies, given a relatively small number of labelled samples, all without model retraining. We previously demonstrated the use of TFM for power grid dynamic security assessment in \cite{arowolo2026revisitingdatadrivendynamicsecurity}. 

In this paper, we propose an approach using TFM as a general-purpose neural solver for power grid steady-state analyses. The proposed approach formulates PF and voltage stability margin (VSM) prediction as tabular regression tasks and N-k contingency screening as a tabular classification task. Our main contributions are as follows:
\begin{itemize}
    \item We propose a TFM approach for power system steady-state analyses. We demonstrate the proposed approach on PF, VSM prediction and N-k contingency screening.
    \item We present simple methods to extract topological features for the proposed TFM approach. The presented methods improve accuracy and scalability to larger grids. 
    \item We demonstrate the accuracy and robustness of our proposed approach through comprehensive case studies on the 118- and 2000-bus grids.
\end{itemize}
The remainder of the paper is structured as follows: Section 2 formulates the relevant problems, Section 3 presents our proposed methodology, Section 4 presents the case studies, Section 5 discusses the results, and Section 6 concludes the paper.

\section{Problem Formulation}
This section provides an overview of the three steady-state analysis tasks: AC PF and its DC approximation, $N-k$ contingency screening, and VSM/loadability margin prediction.

\subsection{AC power flow}
AC PF determines the flow of electrical power in a steady-state power grid, and it is mathematically formulated as a set of non-linear equations. Let $G$ be a power grid graph with $ \mathcal{N}$ buses and $\mathcal{E}$ branches (transmission lines and transformers). The branch status vector $s \in \{0,1\}^{\mathcal{E}}$ represents the set of active branches, and $Y(s) = G(s) + jB(s)$ is the admittance matrix which represents the grid topology. For each bus $i \in \mathcal{N}$, the following four variables are defined: voltage magnitude $V_i$, voltage angle $\theta_i$, active power injection $P_i$, and reactive power injection $Q_i$, where $P_i = P^{\mathrm{gen}} - P^{\mathrm{load}}$ and $Q_i = Q^{\mathrm{gen}} - Q^{\mathrm{load}}$. The AC PF equations are given by:
\begin{align}
P_i &= V_i \sum_{k \in \mathcal{N}} V_k \left( G_{ik} \cos\theta_{ik} + B_{ik} \sin\theta_{ik} \right), \label{eq:acpf-p}\\
Q_i &= V_i \sum_{k \in \mathcal{N}} V_k \left( G_{ik} \sin\theta_{ik} - B_{ik} \cos\theta_{ik} \right), \label{eq:acpf-q}
\end{align}
where $\theta_{ik} = \theta_i - \theta_k$. In PF, each bus has two known variables and two unknown variables, depending on the bus type, as shown in Table \ref{tab:bus-types}. The above leads to a set of $2\mathcal{N}$ equations with $2\mathcal{N}$ unknowns. Classic AC PF solvers use the iterative Newton--Raphson (NR) approach to calculate the unknown voltage variables at PQ and PV buses, while the other variables are analytically derived \cite{rivera_pf_2025}.
\begin{table}[b]
\centering
\caption{Known and unknown variables per bus type.}
\label{tab:bus-types}
\begin{tabular}{lcc}
\toprule
 & \multicolumn{2}{c}{Variable} \\
\cmidrule(lr){2-3}
Bus type & Known & Unknown \\
\midrule
PQ bus    & $P_i,\, Q_i$      & $V_i,\, \theta_i$ \\
PV bus    & $P_i,\, V_i$      & $Q_i,\, \theta_i$ \\
Slack bus & $V_i,\, \theta_i$ & $P_i,\, Q_i$ \\
\bottomrule
\end{tabular}
\end{table}

\subsection{DC Power Flow Approximation}
AC PF in Eqs. \ref{eq:acpf-p}-\ref{eq:acpf-q} has no analytical solution and must be evaluated through NR iterations. For transmission systems, AC PF is often simplified with the following assumptions: i) conductances $G_{ik}$ are zero; ii) phase angle differences $\theta_{ik}$ are small; iii) voltages are constant and set to $1.0$ pu. Eqs. \ref{eq:acpf-p}-\ref{eq:acpf-q} reduce to:
\begin{equation}
P_i = \sum_{k \in \mathcal{N}} B_{ik}\, \theta_{ik}. \label{eq:dcpf}
\end{equation}
DCPF given in Eq. \ref{eq:dcpf} is linear and can be easily solved to obtain the unknown voltage angles at each bus in the grid.

\subsection{AC Power Flow Prediction}
Solving the AC PF equations can be formulated as a regression learning task. The task objective is to predict the unknown $2\mathcal{N}$ variables. The implicit function in Eq. 4 describes the true AC PF solution:
\begin{equation}
F(x, u, s) = 0, \label{eq:solution-map}
\end{equation}
where $x \in \mathcal{X} \subseteq \mathbb{R}^{2\mathcal{N}}$ are the known variables, $u \in \mathcal{U} \subseteq \mathbb{R}^{2N}$ are the unknown variables, and $s \in \{0,1\}^{\mathcal{E}} $ are the active branches. A machine learning model aims to learn an approximate solution
\begin{equation}
\tilde{F}(x, s) \rightarrow u, \label{eq:approx-map}
\end{equation}
where AC PF problem instances are provided in the training data.

\subsection{$N-k$ Contingency Screening}
Let $c$ be a set of contingencies or simultaneous outages of interest, and let $S_i$ be the vector of nodal apparent power injections, $S_i = \sqrt{(P_i)^2 + (Q_i)^2}$. The system operation seeks to know if $S_i$ is feasible for all possible contingencies $c$, $\forall\, 1 \leq c \leq k$. As previously mentioned, $c$ can be very large for $k > 1$ even for a moderately sized grid. Therefore, operators typically rely on $N-1$ contingencies and a few heuristically defined higher-order cases \cite{christianson_fast_2025} to assess steady-state feasibility.
From Eqs. \ref{eq:acpf-p}-\ref{eq:acpf-q}, the active and reactive power flowing on a branch is given by:
\begin{align}
P_{ij} &= |V_i||V_j|\left( G_{ij}\cos\theta_{ij} + B_{ij}\sin\theta_{ij} \right) - G_{ij}|V_i|^2, \label{eq:branch-p}\\
Q_{ij} &= |V_i||V_j|\left( G_{ij}\sin\theta_{ij} - B_{ij}\cos\theta_{ij} \right) + B_{ij}|V_i|^2, \label{eq:branch-q}
\end{align}
$\forall (i,j) \in \mathcal{E}$.
Feasibility is defined by two security criteria. The first is the branch thermal limit defined by
\begin{equation}
S_{ij} < d \cdot R_{ij} \quad \forall (i,j) \in \mathcal{E}, \label{eq:thermal-limit}
\end{equation}
where $R_{ij}$ is the maximum thermal rating of the branch, and $d$ is some allowance factor. The second is the voltage limit:
\begin{equation}
V_i^{\min} \leq V_i \leq V_i^{\max} \quad \forall i \in \mathcal{N}. \label{eq:voltage-limit}
\end{equation}
The prediction task is to determine the feasibility of a set $\{x, s^c\}$, where $s^c$ is the post-contingency topology. The prediction task can be formulated in two ways. The first is to learn a classifier $G : (x, s^c) \rightarrow \hat{y}$, which directly maps the known variables into the contingency label $y \in \{0,1\}$. The second approach uses the learnt AC PF solution map $\tilde{F} : (x, s^c) \rightarrow u$. With all PF variables obtained from the predictions, contingency screening is done by applying the feasibility thresholds in Eqs. \ref{eq:thermal-limit}-\ref{eq:voltage-limit} to the predicted variables.

\subsection{Voltage Stability Margin Prediction}
Voltage stability margin (VSM) represents how close the grid is to voltage collapse and quantifies the maximum additional load the grid can admit before instability occurs. Numerical solvers such as NR are prone to divergence at or close to the point of instability \cite{rivera_pf_2025}, which is also known as the 'nose' of the PV curve. The nose point is conventionally derived numerically through continuation power flow (CPF). CPF is done by scaling all specified injections along a direction while holding voltage setpoints fixed:
\begin{align}
P_i &= \tau P_0, \label{eq:cpf-p}\\
Q_i &= \tau Q_0. \label{eq:cpf-q}
\end{align}
Here we assume proportional scaling of power injections from the base case $P_0$, $Q_0$. For each value of $\tau$, NR is used to trace out the PV curve. At the maximum loading $\tau_{\max}$, $\det J = 0$ for the PF Jacobian (singular Jacobian). The loadability factor $\sigma$ at every point along $1 \leq \tau \leq \tau_{\max}$ is given by
\begin{equation}
\sigma = \frac{\tau_{\max}}{\tau}, \label{eq:loadability}
\end{equation}
and decreases monotonically from the base injection to the nose point where $\sigma = 1$. $\sigma - 1$ is the loadability margin and is the fraction by which every injection in the grid can be scaled before voltage instability is reached. Determining VSM is computationally expensive because CPF involves repeated AC PF solutions. DCPF also cannot accurately determine VSM because the assumptions of DCPF are violated at high loading (stressed grid condition). Hence, we formulate the VSM prediction task as a regression task:
\begin{equation}
H(x, s) \rightarrow \sigma \label{eq:vsm-task}
\end{equation}

\section{Methodology} 

\subsection{Tabular Foundation Model}
A tabular foundation model is a single transformer model pretrained on millions of synthetic datasets sampled from a tabular data-generating prior \cite{muller2022transformers}. The pretraining step is done once during the model development. Afterwards, the pretrained transformer becomes a generic learning algorithm that can be used to predict arbitrary unseen datasets without fine-tuning or retraining \cite{hollmann_tabpfn_2023}. During inference, the TFM carries out in-context learning, where the model receives an unseen dataset with both labelled training samples and unlabelled test samples; the model then simply processes the new input-output pairs in a single forward pass without any weight update. Given a table of labelled examples,
\begin{equation}
\mathcal{C} = \{(z_j, y_j)\}_{j=1}^{N}, \label{eq:context}
\end{equation}
which serve as the model context, and unlabelled rows $z_*$ which are the model query, the TFM predicts labels for all query rows during the forward pass, as shown in Fig. 1. 

ICL by TFM is fundamentally different from standard supervised learning. In supervised learning, the model weight is fitted to each dataset via gradient update:
\begin{equation}
  \theta^{*} = \arg\min_{\theta} \sum_{j=1}^{N}
  \mathcal{L}\big(f_{\theta}(z_j),\, y_j\big),
  \qquad \hat{y}_{*} = f_{\theta^{*}}(z_{*}),
\end{equation} 
This means a new task, grid topology or range of OPs require retraining by solving Eq. 15 again. By contrast, TFM is pretrained on datasets drawn from a prior over
data-generating process:
\begin{equation}
  \phi^{*} = \arg\min_{\phi}\;
  \mathbb{E}_{\substack{\psi \sim p(\psi)\\ (C,\, z_*,\, y_*) \sim p(\cdot \mid \psi)}}
  \big[ -\log q_{\phi}(y_{*} \mid z_{*}, C) \big],
\end{equation}
After pre-training, the TFM performs ICL and predicts labels for any new tabular dataset directly at inference \cite{hollmann_accurate_2025}. Adapting the TFM to a new task, grid topology or range of OPs in power systems only requires changing $\mathcal{C}$, no gradient update is required.
\begin{figure}[t]
    \centering
    \includegraphics[width=0.48\textwidth]{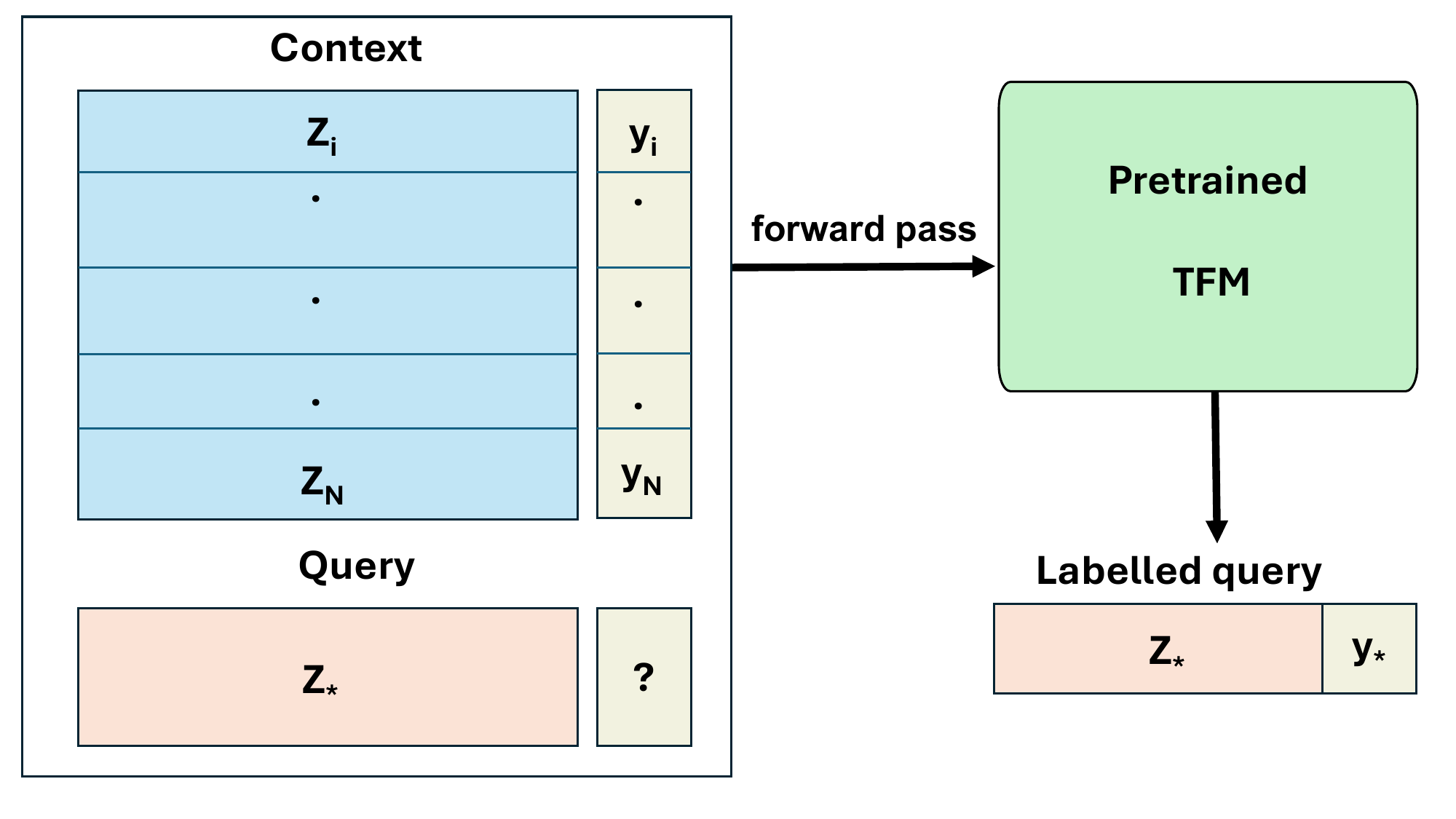}
    \vspace{-4mm}
    \caption{The whole table of labelled context rows and unlabelled query rows is passed to the pretrained TFM in one step. The TFM returns labels for all query rows in a single forward pass.}
    \label{fig:TFM_DSA_workflow}
    % \vspace{-0.1cm}
\end{figure}
\vspace{-0.1cm}
As described in \cite{nagler_statistical_2023}, the TFM approximates the posterior predictive distribution at inference:
\begin{equation}
p(y_* \mid z_*, C) = \int p(y_* \mid z_*, \psi)\, p(\psi \mid C)\, d\psi, \label{eq:ppd}
\end{equation}
where $\psi$ is the latent parameter governing the data generation process of the prior. The prediction step is therefore an approximate Bayesian inference conditioned on the available training data. It is important to stress that no real data is used during pretraining, but the synthetic data comes from a prior of randomly sampled structural causal models (SCM) \cite{hollmann_tabpfn_2023}. We have used \textit{TabPFNv3} as the TFM in this paper \cite{grinsztajn_tabpfn-3_2026}. The TabPFN model uses both row and column attention; therefore, predictions are independent of row or column order. The model also supports classification and regression tasks and returns a predictive distribution for each label rather than a point prediction. 
% To use TabPFN for power grid tasks, the intended task must be organized as a tabular classification or regression. Power grid steady-state analysis tasks therefore have to be converted into tabular prediction. For PF prediction, the operating condition (loading and generator setpoints) becomes the features in the context table, and nodal target variables become independent labels. A similar approach is used for VSM prediction, where the model predicts a single scalar margin per grid instance. The classifier variant of TabPFN is used for the contingency screening task.

\subsection{Predicting power flow residuals}
For TabPFN, each PF example is one row in the tabular data, with the $2\mathcal{N}$ known variables as feature columns, while unknown variables are predicted per column. A table of size $n_{\mathrm{samples}}\times 2\mathcal{N}$ is used to predict each unknown variable in turn. For example, $V_i$ at a given load bus is predicted for all samples in one inference step.  Rather than predicting the voltage state directly, we can predict its residual or the deviation from a linear approximation of the state. DCPF is a fast linear approximation of the AC PF equations: under the DC assumptions, the voltage magnitude at every bus is $1$ pu (the flat profile) and $\theta={B}'^{-1}{p}$, where $B'$ is the DC susceptance matrix Laplacian with the slack row removed as defined in Eq.3. The difference between the DC approximation and the true AC solution captures the nonlinear effects
of the PF problem. We define the residual of the DC solution as:
\begin{equation}\label{eq:residual}
\Delta{V} = {V}_{\mathrm{AC}} - {1},
\qquad
\Delta{\theta} = {\theta}_{\mathrm{AC}} - {B}'^{-1}{p},
\end{equation}
and reformulate the voltage-state prediction tasks as residual learning. The unknown injections are reconstructed from the voltage states.

\subsection{Scaling TFM to larger grids with power decoupling}
The inference cost of TFMs scales with the size of the context table. For example, the TabPFN model we use in this paper is pretrained on tables with at most 2000 feature columns. Larger tables may be processed during inference if they fit into the GPU memory, but this may come at the cost of degraded accuracy. In PF, the number of features in the context table grows as $2\mathcal{N}$. For the 2000-bus system, we have $4000$ feature columns. The original TabPFN model can be used to process all 4000 features. Alternatively, the decoupled power-flow assumption can be applied for feature selection. In transmission networks, voltage angles are strongly coupled with the active-power injections, while voltage magnitudes are strongly coupled to the reactive-power flow. This separates the power-flow equations into two independent half-systems, 
\begin{equation}\label{eq:halfsystems}
\Delta\ \theta = {B}'^{-1}\,\Delta{P},
\qquad
\Delta{V} = {B}''^{-1}\,\Delta{Q},
\end{equation}
where ${B}''$ is the reduced susceptance matrix of the PQ buses. Under the assumption that the $P$--$V$ and $Q$--$\theta$ cross-couplings are weak, TabPFN can approximate ${\theta}$ using only the known ${P}$ as feature columns, and approximates ${V}$ using only ${Q}$ as feature columns.

\subsection{Encoding grid topology in tabular data}
A TFM does not automatically encode a notion of the grid topology. Unlike GNN models, where the grid topology is provided directly to the model as inductive bias, the grid topology can only be specified to a TFM through the data provided as context. The topology may be encoded using one-hot vectors specifying connected and disconnected branches, but this leads to a large increase in the number of features for large grids and also generalizes poorly \cite{qi_continuous_2025}. Here, we propose encoding the grid topology in the TFM using the well-known line outage distribution factors (LODF).

For a grid with $\mathcal{N}$ buses and $\mathcal{E}$ branches, let ${A}\in\mathbb{R}^{\mathcal{E}\times(\mathcal{N}-1)}$ be the reduced branch incidence matrix with rows ${a}_e^{\!\top}$, and let ${B}_D=\mathrm{diag}(1/x_e)$ be the diagonal branch-susceptance matrix, so that ${B}'={A}^{\!\top}{B}_D{A}$ is the Laplacian of the DC susceptance matrix. The power transfer distribution factor (PTDF) matrix is:
\begin{equation}\label{eq:ptdf}
\mathrm{PTDF} = {B}_D\,{A}\,{B}'^{-1},
\end{equation}
whose entry $\mathrm{PTDF}(e,i)$ gives the share of a unit active-power injection at bus $i$ that flows on line $e$. The $\mathrm{LODF}(e,h)$ is the fraction of the flow of line $h$ that is rerouted onto line $e$ when line $h$ is disconnected, and is related to the PTDF through the
line-to-line sensitivities ${\Psi}=\mathrm{PTDF}\,{A}^{\!\top}$:
\begin{equation}\label{eq:lodf}
\mathrm{LODF}(e,h) = \frac{\Psi_{eh}}{1-\Psi_{hh}}.
\end{equation}
We propose a topology feature that combines the LODF with the pre-outage active power flows,
\begin{equation}\label{eq:topofeat}
\Delta{\theta}(s,{p}) =
\bigl({B}'(s)^{-1}-{B}'^{-1}\bigr){p},
\end{equation}
one column per non-slack bus. For an $N\!-\!k$ contingency, the $N\!-\!k$ LODF matrix is obtained by a Woodbury matrix update \cite{cremer_polynomial_2025}:
with ${U}$ the incidence columns of the outaged branches and ${B}_D^{s}$ the diagonal of ${B}_D$ on the outaged set $s$,
\begin{equation}\label{eq:bprime-s}
{B}'(s) = {B}' - {U}{B}_D^{s}{U}^{\!\top},
\end{equation}
\begin{equation}\label{eq:woodbury}
\begin{split}
{B}'(s)^{-1} = {B}'^{-1} + {B}'^{-1}{U}
\bigl(({B}_D^{s})^{-1}&-{U}^{\!\top}{B}'^{-1}{U}\bigr)^{-1}\\
&\times\,{U}^{\!\top}{B}'^{-1}.
\end{split}
\end{equation}
The inverse is fast to compute for low-rank changes to ${B}'$, such as $N\!-\!1$ up to $N\!-\!3$ contingencies. The proposed topology feature $\Delta{\theta}$ adds power-grid inductive bias to the TFM.

\section{Case studies}
\subsection{Dataset}
We use the  PF$\Delta$ \cite{rivera_pf_2025} PF benchmark dataset in our experiments. The
dataset consists of solved AC PF instances across various grids. The dataset also includes separate subsets to evaluate generalization to $N$-1 and $N$-2 component outages as well as generalization to close-to-infeasible (C2I) instances. The C2I instances are obtained by using \textit{MATPOWER CPF}. We specifically use the C2I full topology data subset for the VSM prediction task. The CPF starts from a base OP, and all loads and injections are proportionally scaled up to 2.5$\times$ the base values. The 118-bus grid has 3600 training samples and 200 testing samples, while the 2000-bus grid has 600 training samples and 92 testing samples. PF$\Delta$ provides the base OP and nose OP for each training/testing sample. The samples have a 'family' index to connect each base OP to its nose OP. Furthermore, we generate 8 OPs for each family according to the following setting: $\tau = 1 + f(\tau_{max} - 1) \forall f \in \{0.3, 0.5, 0.7, 0.8, 0.9, 0.95, 0.97, 0.99\}$. We use the NR solver at each $\tau$ by warm-starting from the previous $\tau$ in the OP family. We call these samples augmented OPs. We label every augmented OP with its proportional loadability margin $\sigma = \tau_{max}/ \tau$.

In addition to the original PF$\Delta$ dataset, we generate additional data for the contingency screening task using the GridFM datakit tool. Because of the use of a custom, relaxed ACOPF formulation in the PF$\Delta$ dataset, all the PF instances violate the steady-state operating limits \cite{puech_gridfm-datakit-v1_2025}. On the 118-bus and 2000-bus grids we consider in this work, all samples were outside the steady-state voltage or branch power operating limits. This makes the dataset alone unsuitable for predicting steady-state violations. The GridFM datakit dataset, however, has a more realistic balance between samples within and outside normal operating limits. We generated 1,558,730 $N$-1 scenarios by disconnecting each branch across 8,760 load scenarios. About 64\% of the $N$-1 scenarios are secure. We similarly generated 3,133,892
$N$-2 scenarios by disconnecting each combination of two branches across 200 load scenarios.
About 41\% of the $N$-2 scenarios are secure. Scenarios which lead to grid islanding are
discarded.

\subsection{Experimental setting} 
The case studies for the PF and the VSM prediction tasks are on the 118-bus and 2000-bus grids. Only the 118-bus grid was used for the contingency screening task because the problem becomes intractably large for the 2000-bus system. We use a wide range of baseline models in the experiments. We have kept the same baseline models across the 3 tasks for consistency in our analysis. The baseline models are:

\begin{enumerate}
\item Graph Neural Solver (GNS) \cite{donon_neural_2020}: This is a self-supervised GNN approach which predicts nodal voltage angles and magnitudes by minimizing the power imbalance in the network. As such, the model does not require ground truth PF solutions. Generator setpoints are analytically derived from the predicted voltages. GNS uses the following hyperparameters: $k = 10$, hidden dimension $= 64$, $\gamma = 0.01$.
\item Typed-GNN (TGN) \cite{lopez-garcia_power_2023}: This is also a self-supervised GNN approach where nodal voltages and angles are determined by minimizing power imbalances while deriving generator injections from predicted voltages. TGN defines branches as node types such that different node types can have different update functions and output dimensions. We use TGN with 15 layers, 3 message passing steps and a hidden dimension of 16 per node.
\item PowerFlowNet \cite{lin_powerflownet_2024}: This is a supervised learning GNN approach where all known PF variables are predicted by minimizing the L2 loss between the predictions and the ground truth. We use PowerFlowNet with $k = 3$ and 4 TAGConv layers.
\item Physics-Informed Edge Varying GNN (PI-EVGNN) \cite{nakiganda_graph_2025}: An edge varying GNN differs from standard graph convolution by allowing different trainable weights per node and edge for each convolution layer. It uses a semi-supervised GNN approach which combines the L2 loss between the model predictions and ground truth with a physics-informed power balance loss. We use a
weighting factor ($\lambda = 0.1$) for the physics loss for training stability, with the
following parameters: $k = 3$, number of layers $= 2$, hidden dimension $= 128$.
\item LeapNet \cite{donon_leap_2020}: This is a supervised learning approach which uses standard MLPs. LeapNet encodes the grid topology through a learnable vector $\rightthreetimes$. On the reference, full-topology dataset, $\rightthreetimes = 0$. We use LeapNet with 2 layers and a hidden dimension of 128.
\end{enumerate}

The TabPFN model uses the training split of each experiment as its context and the test split as its query. The same context is used for all queries in experiments where inference is done in batches. Model hyperparameters are selected to be consistent with the literature. For the VSM prediction and $N$-$k$ contingency screening tasks, the nodal outputs of the graph models are aggregated through an MLP readout layer with a hidden dimension of 64 and an output layer of 1. All baseline models were trained with a learning rate of $10^{-4}$, a batch size of 256, the AdamW optimizer and the OneCycleLR learning rate scheduler over 200 epochs. All experiments were repeated with 3 random seeds (3, 7, 11), with the average results reported. The seed changes the sampled training/test data. All training and inference experiments were performed on a computer with an AMD EPYC-7302 CPU with 64 cores and an NVIDIA H100 GPU with 128\,GB memory.

\subsection{Power flow analysis}
\begin{figure*}[t]
  \centering
  \includegraphics[width=\textwidth]{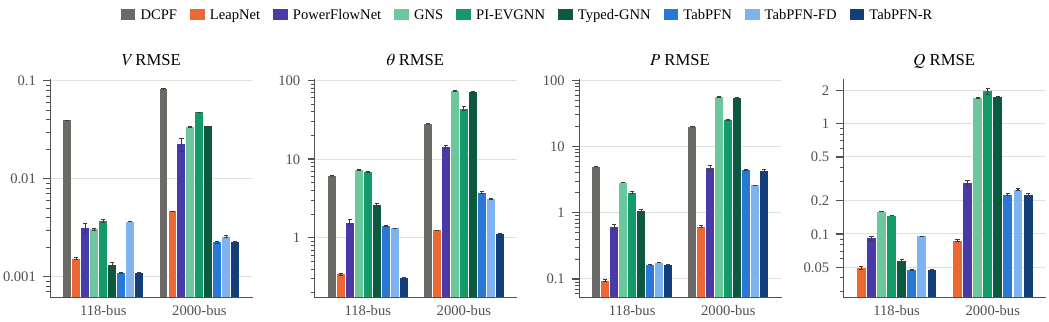}
  \caption{In-distribution accuracy on the full-topology test sets: RMSE of voltage magnitude
  $V$, voltage angle $\theta$, active power $P$ and reactive power $Q$ (DCPF does not output
  $Q$). Bars show the mean and error bars the standard deviation over three seeds in log scale.}
  \label{fig:id}
\end{figure*}

\begin{figure*}[t]
  \centering
  \includegraphics[width=\textwidth]{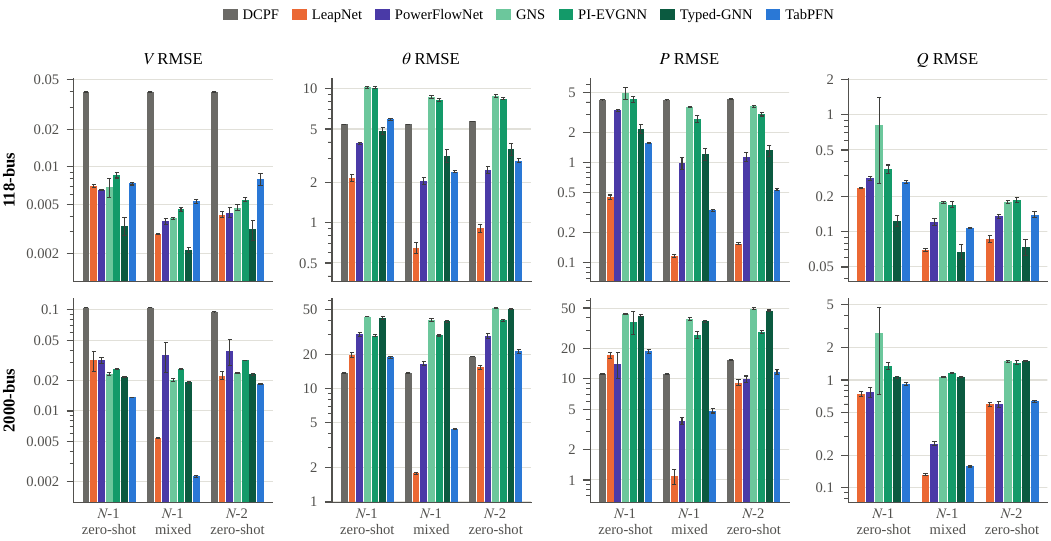}
  \caption{Generalization to unseen line outages on the 118-bus (top) and 2000-bus (bottom)
  systems. Zero-shot: outage topologies absent from training (and from TabPFN's context);
  mixed: $N$-1 samples included. Mean and standard deviation over three seeds in log scale;
  DCPF does not output $Q$.}
  \label{fig:topo}
\end{figure*}

\begin{figure*}[t]
  \centering
  \includegraphics[width=\textwidth]{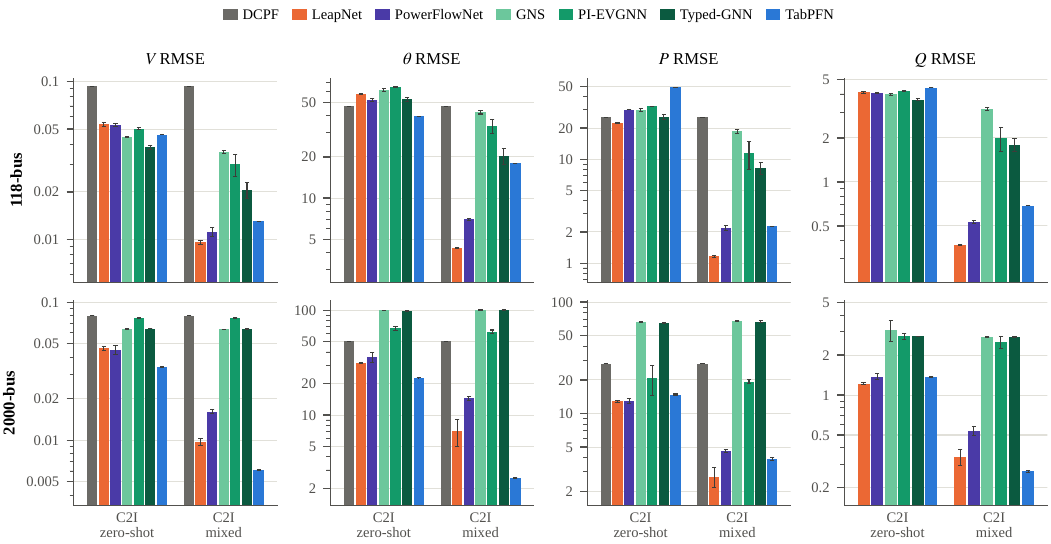}
  \caption{Generalization to close-to-infeasible (C2I) operating points on the 118-bus (top)
  and 2000-bus (bottom) systems. Zero-shot: C2I samples absent from training (and from
  TabPFN's context); mixed: C2I samples included. Mean and standard deviation over three
  seeds (log scale); DCPF does not output $Q$.}
  \label{fig:c2i}
\end{figure*}

\begin{figure*}[t]
  \centering
  \includegraphics[width=\textwidth]{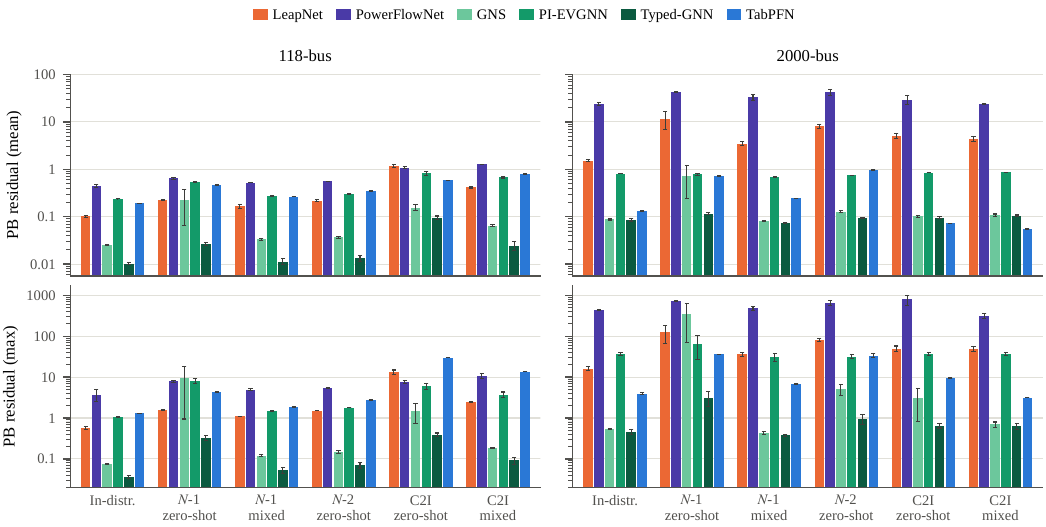}
  \caption{Mean (top) and maximum (bottom) power-balance residual of the learned models across
  all test conditions: We report the mean and standard deviation over three seeds (log
  scale).}
  \label{fig:pb}
\end{figure*}
The full topology dataset has 56000 samples for the 118-bus and 28000 samples for the 2000-bus grids. We split the data into 80\% for training, 10\% for validation and 10\% for testing.  Fig. 2 compares TabPFN to the baseline approaches for the in-distribution generalization
task on the full topology datasets. On the 118-bus grid, TabPFN has the lowest error at
predicting voltage magnitudes and reactive power injection of all models, with $V$ RMSE of
0.0011 and $Q$ RMSE of 0.047. TabPFN also has the second-lowest error to LeapNet in predicting
slack bus active power injection $P$ and voltage angles $\theta$. TabPFN attains lower errors
than all 4 GNN-based baselines across all 4 predicted variables. It is also notable that
relative to DCPF, TabPFN's RMSE is $36\times$ lower for $V$, $30\times$ lower for $P$, and
$4.4\times$ lower for $\theta$. On the 2000-bus grid, all the models record higher errors.
TabPFN maintains the lowest $V$ error, with errors lower than the GNN baselines across all the
predicted variables. We also see that the physics-informed GNN models have similar errors to the DCPF method. Further, we assess two TabPFN variants: the residual variant (TabPFN-R)
where the model predicts the DCPF residual of $V$ and $\theta$ according to Eq. 16, and
the fast-decoupled variants where $(V, Q)$ and $(\theta, P)$ are predicted separately
according to Eq. 17. TabPFN-R has the lowest error for 3 of the 4 predicted variables on
the 118-bus grid and the lowest $V$ and $\theta$ errors on the 2000-bus grid. This suggests that
residual prediction is an effective way to further improve the accuracy of TabPFN on the task.
Also, TabPFN-FD has slightly lower errors for $\theta$ and $P$ prediction errors on the 2000-bus system and similar errors to the original model for $V$ and $Q$. This highlights how TabPFN may be scaled to larger systems, via similar decomposition, without loss of accuracy.

We perform 3 topology generalization experiments as shown in Fig. 3. The first is `$N$-1 zero-shot' where the models are trained with only the full topology training set and evaluated on the $N$-1 test set. The second is `$N$-1 mixed' where the models are trained with a mix of 50\% of the full topology training set and the whole $N$-1 data training set. We evaluate the models on the $N$-1 test set. The third experiment uses the same training data as in the second experiment, but models are evaluated zero-shot on the $N$-2 data test set ($N$-2 zero-shot). 
On both test grids, the pretrained TabPFN has comparable errors to the trained GNN baselines. The performance of TabPFN is especially strong on the 2000-bus grid, where the model achieves the lowest or second-lowest errors for all 4 predicted variables across the 3 experiments. This is in contrast to the GNN models, which have substantially larger errors on the 2000-bus grid, which points to a structural limitation of message-passing architectures on larger grids. In the zero-shot experiments, TabPFN has comparable $\theta$ and $P$ errors to DCPF. However, the use of N-1 samples in its context in the mixed experiment leads to lower errors compared to DCPF. The GNN models also have worse errors in the mixed experiment compared to TabPFN and the MLP-based LeapNet. Overall, zero-shot generalization remains a weakness for the ML models, but TabPFN can adjust its prediction to new topologies via ICL alone, without retraining.  

Next, we investigate generalization to close-to-infeasible (C2I) PF cases with two experiments as shown in Fig. 4. In the `C2I zero-shot' experiment, the ML models are trained with
the feasible full topology data train set and evaluated on the C2I data test set. In the
`C2I mixed' experiment, the ML models are trained with a mix of 50\% of the feasible full
topology data train set, the C2I data train set, and the 'approaching infeasibility' data train
set from \cite{rivera_pf_2025}. The trained models are evaluated on the C2I data test set. 
On the C2I zero-shot experiment, TabPFN has lower $V$ and $\theta$ errors for both test grids, compared to DCPF, which is known to be unsuitable for C2I PF cases. However, TabPFN has a larger $P$ error. Once again, TabPFN has the best performance on the 2000-bus grid, with the lowest $V$, $\theta$, and $Q$ errors across both zero-shot and mixed experiments. Notably, TabPFN errors are significantly reduced in the mixed experiment compared to the zero-shot experiment, while the GNN models trained with physics loss maintain consistently high error levels in both experiments. The superior accuracy of TabPFN on the large test grid and its adaptivity to samples from different PF scenarios via ICL highlight its potential for PF analysis.

While RMSE and MSE typically measure the prediction errors of ML models in PF analysis, predictions must also be physically consistent and satisfy the power balance (PB). PB residual measures the deviation from the expected condition of zero power imbalance. Fig. 5 shows the mean and maximum PB residual as in \cite{rivera_pf_2025} for all experiments. We observe that the type of loss function employed during training impacts PB residuals. PowerFlowNet and LeapNet, which are trained on supervised loss only, have comparable residuals to TabPFN on the 118-bus grid and higher residuals on the 2000-bus grid. GNS and TGN, which are trained in a self-supervised fashion specifically to minimize PB residuals, have consistently lower PB errors than TabPFN. While training with PB loss improves the physical consistency of model predictions, the predictions may be physically consistent but still have high errors compared to the ground truth solutions. For example, GNS and TGN have larger RMSE errors than TabPFN in most experiments despite having lower PB residual. Since a low PB residual implies the predicted voltages and injections are physically consistent,
not that they match the true operating point, PB residuals and predicted variable errors
should be reported and assessed together.

\subsection{N-k contingency screening}
\begin{figure*}[t]
  \centering
  \includegraphics[width=\textwidth]{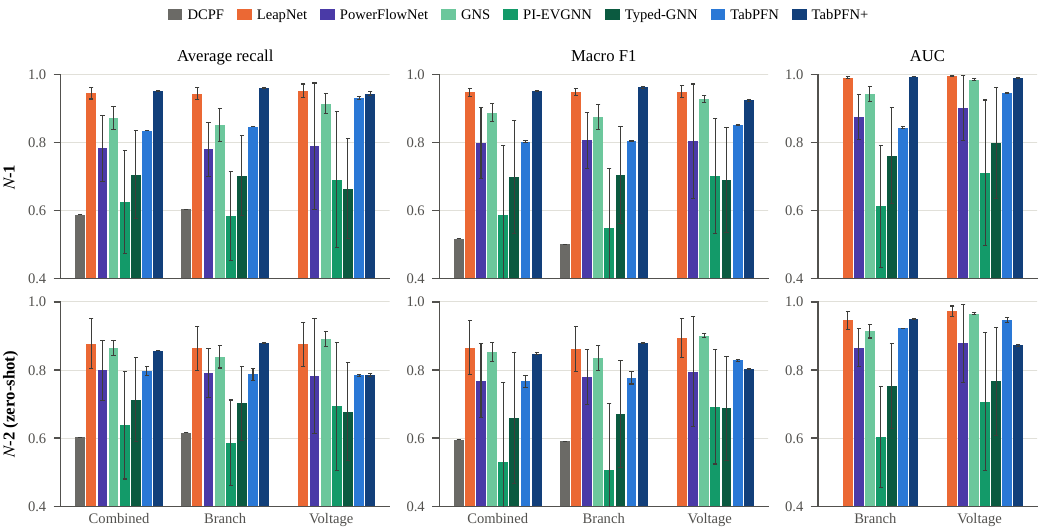}
  \caption{N-k contingency classification on the 118-bus grid for $N$-1 contingencies (top)
  and unseen $N$-2 contingencies (bottom): average recall, macro F1 and AUC for the
  combined, branch and voltage violation prediction. We report mean and standard deviation over three seeds. DCPF cannot be used to assess voltage violations.}
  \label{fig:nk_direct}
\end{figure*}

\begin{figure*}[t]
  \centering
  \includegraphics[width=\textwidth]{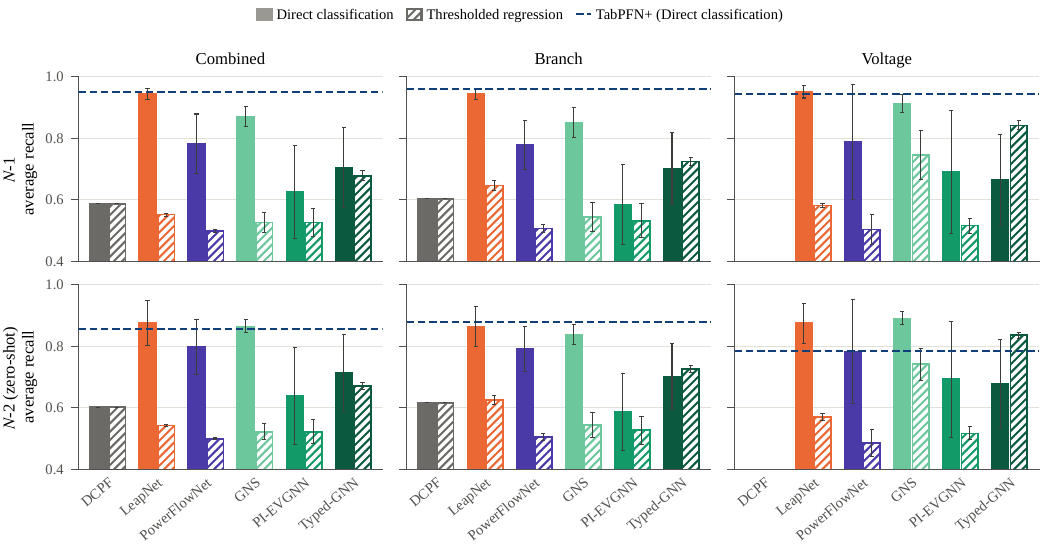}
  \caption{Average recall of contingency screening obtained by thresholding the PF regression
  outputs (hatched) compared with the directly trained classifiers (solid). The dashed line
  shows the performance of the direct TabPFN+ classifier. We report mean and standard deviation over three seeds.}
  \label{fig:nk_screen}
\end{figure*}

In this study, we compare the baseline models trained as classifiers with the TabPFN model. We randomly sample 10000 full topology samples from
PF$\Delta$ (with violations) and 10000 from the GridFM data, split 80\% and 20\% for training (context) and validation sets. In addition, 10000 randomly selected N-1 samples are added to the training data. No N-2 samples are seen during training. Each baseline model is trained as a multi-label classifier. For the GNN-based models, the aggregated graph embeddings are passed into an MLP prediction head with 2 output neurons, one for each violation type (bus voltage and branch violations). \texttt{BCEWithLogitsLoss} is used as a loss function to output the probability of branch/voltage violations independently. At inference, each output neuron is independently thresholded with sigmoid $> 0.5$ indicating a violation. TabPFN classifier only predicts one label, so we use different classifiers for voltage and branch violation screening. We show the performance of the models for the combined violation set and for voltage and branch violations separately in Fig. 6. We make the following observations: First, the topology-augmented TabPFN model (TabPFN+) has strong screening performance. On the held-out $N$-1 dataset, it had a combined screening average recall of 0.95 and 0.86 on the unseen $N$-2 dataset. The TabPFN+ model also had better screening accuracy than the base TabPFN model, which lacked topology features. This highlights the importance of informative engineered features for ICL in this application since the model is more constrained in learning only from the provided context samples. TabPFN+ is also better at screening branch violations than voltage violations. We hypothesize that this may be due to the use of linear change in bus voltage angles as topology features, as shown in Eq. 20, since the topology feature augments the model with knowledge of how power flows redistribute after a branch outage.

Second, the ML classifiers were much better than DCPF at screening branch violations. The base TabPFN model achieved an average recall of 0.85 for branch violations compared to DCPF with 0.60 for held-out $N$-1 contingencies and 0.79 for $N$-2 contingencies compared to DCPF with 0.62. All the trained classifiers perform better than DCPF at contingency screening except for PI-EVGNN. DCPF is also unable to screen voltage violations since voltages are assumed to be 1 p.u.\ at every
bus. On the other hand, the neural surrogates excel at screening voltage violations with AUC of 0.99 and 0.95 for LeapNet and TabPFN, respectively, on held-out $N$-1 contingencies and 0.97 and 0.95 on millions of unseen $N$-2 contingencies.

Third, the base TabPFN model without topology-specific features shows strong generalization to unseen $N$-2 topologies. Using the threshold-free AUC metric, TFM performs on par with the best trained classifier models (LeapNet and GNS) on $N$-2 topologies, which are never present in its context. The TabPFN model has an AUC of 0.92 for branch violation screening on $N$-2 contingencies compared to 0.91 and 0.95 for GNS and LeapNet, respectively. It also has an AUC of 0.95 for voltage violation screening compared to 0.97 and 0.97 for LeapNet and GNS. This shows that TabPFN can separate $N$-2 scenarios which lead to violations and those scenarios which do not, only through ICL. However, the lower average recall (0.80) suggests that a classification threshold of 0.5 is not optimal.

Finally, direct classification for contingency screening is much better than screening with a PF neural surrogate. With a PF surrogate, voltage setpoints are predicted for each contingency sample. The predicted voltage magnitudes can be compared to the voltage limits, while the branch flows can be derived from the predicted voltage angles according to Eqs. 6-7 and compared to branch thermal limits as well. This predict-then-threshold approach is used in \cite{nakiganda_graph_2025} and is how the DCPF screening is performed. We compare the DCPF baseline to other model architectures when they are used as PF surrogates in Fig. 7. 
All the PF neural surrogates had much worse screening accuracy compared to their direct classifiers. Also, only TGN outperforms DCPF at combined branch and voltage violation screening on both datasets. Since contingency screening relies on accurately determining extreme errors in branch angles or voltage magnitudes, the poor screening accuracy of the predict-then-threshold approach of PF neural surrogates can be attributed to the smoothing of extreme errors induced by the MSE loss used for training. 

\subsection{Voltage stability margin prediction}

\begin{figure*}[t]
  \centering
  \includegraphics[width=\textwidth]{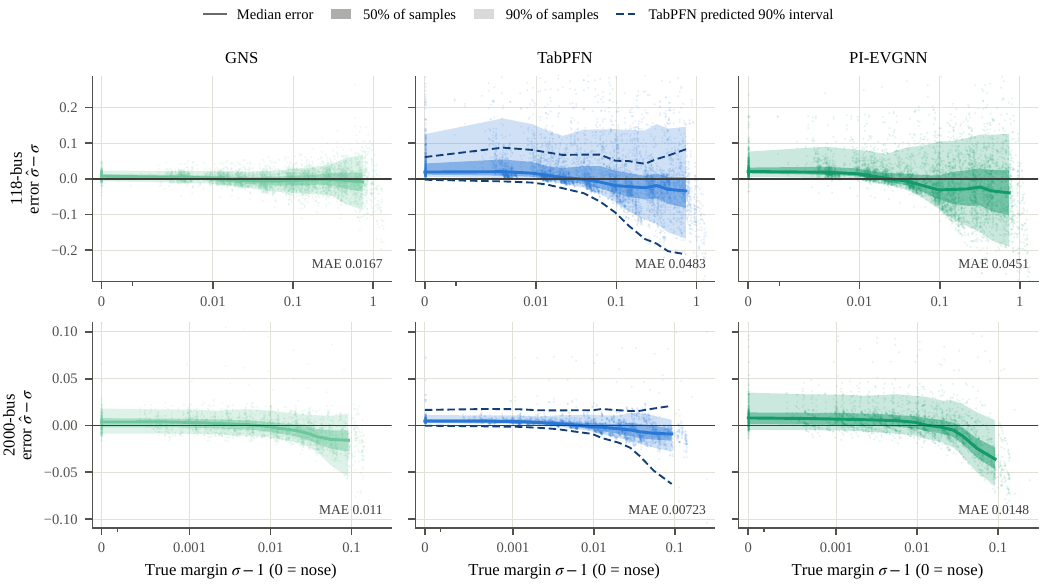}
  \caption{Prediction error of the loadability factor, $\hat{\sigma}-\sigma$, against the true
  margin $\sigma-1$ in the in-distribution experiment on the 118-bus (top) and 2000-bus (bottom) grids. The solid line is the median error, and the shaded bands contain 50\% and 90\% of the test points. The dashed lines show the predicted 90\% interval of TabPFN.}
  \label{fig:vsm_margin}
\end{figure*}

In this study, we predict the loadability factor $\sigma$ in two experiments. The VSM data is split by the family index so no family of OPs appear in both training and testing subset. Each experiment uses known PF variables as input features and predicts $\sigma$ directly without PF solutions. The first experiment assesses the models' in-distribution accuracy and sample efficiency. The in-distribution experiment uses one OP per family, drawn randomly from the base, augmented and nose OPs. The second experiment assesses out-of-distribution generalization of the models. The models are trained with only the base OPs. Base OPs and nose OPs have different distributions. For example, base OPs on the 118-bus grid have 
$\sigma \in [1.11, 2.36]$ at a total loading of 59-65p.u., while the nose OPs have $\sigma = 1.0$ and a total loading of 70-152 p.u. Both experiments have similar training data sizes. (3600 and 600 independent OPs for 118- and 2000-bus systems respectively.) We evaluate the models on all test samples in both experiments.

% ---- ID experiments ------------------------------------------------------------
\begin{table}[t]
  \centering
  \small
  \setlength{\tabcolsep}{2.5pt}
  \caption{Percentage of VSM predictions within specified tolerance for in-distribution generalization experiment: $|\hat{\sigma}-\sigma| \le 0.01$ for true margins $\sigma-1 \le 0.05$ and $|\hat{\sigma}-\sigma| \le 0.05$ for $\sigma-1 > 0.05.$ We report mean $\pm$ standard deviation over three seeds; $n$ is the number of test samples. Best value per column in bold.}
  \label{tab:vsm_tolerance_id}
  \begin{tabular}{lcccc}
    \toprule
    & \multicolumn{2}{c}{118-bus} & \multicolumn{2}{c}{2000-bus} \\
    \cmidrule(lr){2-3} \cmidrule(lr){4-5}
    True margin $\sigma-1$ & $\le 0.05$ & $> 0.05$ & $\le 0.05$ & $> 0.05$ \\
    $|\hat{\sigma}-\sigma|$ & $\le 0.01$ & $\le 0.05$ & $\le 0.01$ & $\le 0.05$  \\
    & ($n=978$) & ($n=1022$) & ($n=797$) & ($n=123$) \\
    \midrule
    GNS & \textbf{57.7\,$\pm$\,1.2} & \textbf{90.4\,$\pm$\,1.3} & 69.6\,$\pm$\,0.8 & 94.6\,$\pm$\,2.0 \\
    TabPFN & 33.6\,$\pm$\,0.4 & 57.1\,$\pm$\,0.4 & \textbf{83.8\,$\pm$\,1.0} & 96.5\,$\pm$\,1.2 \\
    PI-EVGNN & 30.2\,$\pm$\,0.6 & 49.0\,$\pm$\,2.3 & 60.1\,$\pm$\,4.5 & 76.4\,$\pm$\,11.0 \\
    LeapNet & 34.3\,$\pm$\,0.3 & 44.0\,$\pm$\,1.0 & 77.2\,$\pm$\,3.2 & \textbf{96.7\,$\pm$\,0.8} \\
    PowerFlowNet & 29.0\,$\pm$\,5.9 & 45.8\,$\pm$\,3.9 & 36.3\,$\pm$\,12.3 & 49.3\,$\pm$\,1.2 \\
    Typed-GNN & 13.8\,$\pm$\,3.1 & 30.1\,$\pm$\,4.0 & 32.7\,$\pm$\,6.0 & 54.7\,$\pm$\,15.3 \\
    \bottomrule
  \end{tabular}
\end{table}

% ---- BASE experiments ------------------------------------------------------------
\begin{table}[t]
  \centering
  \small
  \setlength{\tabcolsep}{2.5pt}
  \caption{Percentage of VSM predictions within specified tolerance for out-of-distribution generalization experiment:  $|\hat{\sigma}-\sigma| \le 0.01$ for true margins $\sigma-1 \le 0.05$ and $|\hat{\sigma}-\sigma| \le 0.05$ for $\sigma-1 > 0.05$ We report mean $\pm$ standard deviation over three seeds; $n$ is the number of test samples. Best value per column in bold.}
  \label{tab:vsm_tolerance_base}
  \begin{tabular}{lcccc}
    \toprule
    & \multicolumn{2}{c}{118-bus} & \multicolumn{2}{c}{2000-bus} \\
    \cmidrule(lr){2-3} \cmidrule(lr){4-5}
    True margin $\sigma-1$ & $\le 0.05$ & $> 0.05$ & $\le 0.05$ & $> 0.05$ \\
    $|\hat{\sigma}-\sigma|$ & $\le 0.01$ & $\le 0.05$ & $\le 0.01$ & $\le 0.05$  \\
    & ($n=978$) & ($n=1022$) & ($n=797$) & ($n=123$) \\
    \midrule
    GNS & \textbf{4.7\,$\pm$\,0.7} & \textbf{40.8\,$\pm$\,7.6} & 5.0\,$\pm$\,0.1 & 69.1\,$\pm$\,4.5 \\
    TabPFN & 0.0\,$\pm$\,0.0 & 12.8\,$\pm$\,0.2 & 3.2\,$\pm$\,0.4 & 75.9\,$\pm$\,0.5 \\
    PI-EVGNN & 0.0\,$\pm$\,0.0 & 20.7\,$\pm$\,3.1 & 5.2\,$\pm$\,0.6 & 75.6\,$\pm$\,4.5 \\
    LeapNet & 0.3\,$\pm$\,0.3 & 10.9\,$\pm$\,0.6 & \textbf{6.3\,$\pm$\,1.1} & 72.1\,$\pm$\,2.0 \\
    PowerFlowNet & 0.1\,$\pm$\,0.2 & 16.3\,$\pm$\,1.2 & 2.9\,$\pm$\,3.3 & 79.4\,$\pm$\,4.5 \\
    Typed-GNN & 0.0\,$\pm$\,0.0 & 5.4\,$\pm$\,0.8 & 2.2\,$\pm$\,1.9 & \textbf{79.9\,$\pm$\,5.0} \\
    \bottomrule
  \end{tabular}
\end{table}

Tables II and III compare TabPFN to the baseline models in the in-distribution and out-of-distribution experiments, respectively. We report the share of predictions with absolute error
$|\hat{\sigma}-\sigma| \le 0.01$ for true margins $\sigma-1 \le 0.05$
and $|\hat{\sigma}-\sigma| \le 0.05$ for larger margins. The selected low-margin range of $\sigma - 1$  $\leq 0.05$ is motivated by the WSCC-recommended voltage stability margin in \cite{abed_wscc_1999}. For the in-distribution experiment shown in Table II, GNS is the most accurate model for the 118-bus grid in the low margin range, with 57.7\% of its predictions within the $\pm 0.01$ tolerance of the true margin. TabPFN is second to GNS in the high margin range and on par with LeapNet in the low margin range. On the 2000-bus grid, TabPFN is the most accurate in the low margin range, with similar prediction accuracy to LeapNet in the high margin range. For the out-of-distribution experiments shown in Table III, we see that all the models have poor accuracy in the low margin range, including TabPFN. GNS is the best model on 118-bus grid but only has 4.7\% of predictions within the $\pm 0.01$ tolerance threshold in the low margin range. This once again shows the limitations of the models at extrapolating to previously unseen loading levels and highlights the importance of low margin/near-infeasible OPs in the training data to accurately predict VSM at or near the nose point.

Furthermore, Fig. 8 shows the error distribution of the top 3 models on both test grids for the
in-distribution experiment. For all models, the errors are smallest close to the nose point and
larger with increasing margin. This is desirable from a power system security standpoint, but
also reflects the distribution of the training data. Note that the models overestimate the margin
near the nose, which may be of security concern. In addition to the point prediction, TabPFN also provides a predictive distribution for each prediction. The 90\% quantile interval of its predictive distribution is shown in Fig. 8. We see that the lower bound of the 90\% quantile interval is a conservative estimate for the margin at the nose point, with 88\% and 96\% of samples within the interval on the 118-bus and 2000-bus grids, respectively. This highlights the potential of TabPFN for probabilistic VSM prediction.

\section{Discussions}

The proposed TFM approach uses the pretrained TabPFN model without task-specific fine-tuning, and we find that its performance is competitive with the baseline models, which are always retrained for every task and experiment. We only need to change the model's context for a new experiment
in the case of TFM. Across the three evaluated tasks, no single model was consistently ranked as the best performing. For example, on the 118-bus grid, TabPFN had the lowest prediction error for $V$ and $Q$ for the in-distribution power-flow experiment. However, LeapNet had lower $P$ and $\theta$ errors, while GNS had the most accurate VSM prediction. On the 2000-bus grid, TabPFN consistently ranked near the top, with lower power-flow prediction errors and VSM prediction accuracy. Notably, GNN-based models have larger errors on the 2000-bus system, which
highlights the potential of TabPFN to scale to large systems better than GNN-based models.

All the ML models remain brittle in extrapolation scenarios, including the TFM. In zero-shot experiments for PF and VSM prediction, the performance of the TabPFN models was significantly degraded. However, TabPFN holds an advantage when it comes to generalization because it does not require model retraining for generalization, unlike the baselines. Only a new context is required. The model accuracy increased by about $80\%$ for the VSM task on the 2000-bus system when the context data is changed from base OPs alone to in-distribution OPs. The baselines are unable to match the improved
performance even with retraining with the in-distribution OPs. The proposed TabPFN variants further demonstrate that the inductive bias for model prediction can be added via hand-crafted features and/or targets, instead of via the model architecture, as in GNNs. Predicting DCPF residuals (TabPFN-R) leads to improved accuracy, while decoupling $P\theta$ and $VQ$ enables scaling to the 2000-bus grid without sacrificing accuracy. The proposed linear topology feature also improves the average recall in contingency classification. We note that there is further scope for improved feature
engineering.

Finally, Table~\ref{tab:runtime} compares the computational cost of
the models. TabPFN requires no training, but its inference is slower than that of the baseline models. For contingency screening, TabPFN assesses 3.1 million contingencies in about 2752s, compared to 637s for DCPF, while for PF, its inference is considerably slower because each target variable is predicted sequentially. In theory, each target can be predicted in parallel since they are predicted independently. However, this may rapidly increase memory consumption. The computational benefit of the TabPFN model therefore lies in not requiring training and not in its inference speed.
\begin{table}[t]
\centering
\caption{Average computational cost in seconds: total training time ($tr.$) and total inference time on the test set ($inf.$); $n$ is the number of test
samples. TabPFN requires no training; its context is processed at
inference and included in $inf.$}
\label{tab:runtime}
\footnotesize
\setlength{\tabcolsep}{3.5pt}
\begin{tabular}{lrrrrrr}
\toprule
 & \multicolumn{2}{c}{PF 118-bus} & \multicolumn{2}{c}{PF 2000-bus} & \multicolumn{2}{c}{N-2 screening} \\
 & \multicolumn{2}{c}{($n = 5{,}600$)} & \multicolumn{2}{c}{($n = 2{,}800$)} & \multicolumn{2}{c}{($n = 3{,}133{,}892$)} \\
\cmidrule(lr){2-3}\cmidrule(lr){4-5}\cmidrule(lr){6-7}
Model & $tr$ & $inf$ & $tr$ & $inf$ & $tr$ & $inf$ \\
\midrule
DCPF         & --      & 10      & --       & 10       & --      & 637 \\
LeapNet      & 164     & 11      & 140      & 11       & 101     & 38 \\
PowerFlowNet & 829     & 8       & 4{,}369  & 10       & 512     & 176 \\
Typed-GNN    & 7{,}291 & 10      & 9{,}944  & 12       & 4{,}392 & 621 \\
GNS          & 1{,}901 & 12      & 4{,}065 & 13       & 914 & 177 \\
PI-EVGNN     & 614     & 7       & 2{,}197  & 10       & 437     & 163 \\
TabPFN       & --      & 4{,}731 & --       & 72{,}577 & --      & 2{,}752 \\
\bottomrule
\end{tabular}
\end{table}

\section{Conclusion}
This paper investigated the use of a pretrained TFM (TabPFN) as a unified model for power system steady-state analysis tasks. The proposed model does not require re-training or fine-tuning, but performs ICL on the provided context samples during inference. Through case studies on the 118- and 2000-bus grids, the paper evaluated the accuracy and generalization performance of the proposed approach on PF, VSM prediction and N-k contingency screening. The proposed TFM method is competitive with task-specific baselines across the three tasks and is especially accurate for the 2000-bus PF and VSM tasks. This shows the potential for pretrained TFM in power system workflows. Future work will investigate the scaling up of TFM to even larger test systems, methods to select the best context for ICL, extraction of task-relevant features and computationally efficient inference methods for tasks with multiple outputs such as PF. Overall, our results suggest that pretrained TFMs can serve as a practical and simple alternative to task-specific neural surrogates for power-system steady-state analysis.

% \section*{Acknowledgments}
% This should be a simple paragraph before the References to thank those individuals and institutions who have supported your work on this article.

\bibliographystyle{IEEEtran}

\bibliography{references-2}

% {\appendix[Appendix A]
% Use }

% \vfill

\end{document}